\documentclass[12pt]{article}

\usepackage{scicite}

\usepackage{times}

\usepackage{graphicx}
\usepackage[font=small,margin=12pt,labelfont=bf]{caption}

\newenvironment{sciabstract}{%
\begin{quote} \bf}
{\end{quote}}

\title{Creativity in Robot Manipulation with Deep Reinforcement Learning}

\author
{Juan Carlos Vargas,$^{1}$ Malhar Bhoite,$^{1}$ Amir Barati Farimani$^{1}{}^,{}^{2\ast}$\\
\\
\normalsize{$^{1}$Department of Mechanical Engineering, Carnegie Mellon University, Pittsburgh PA 15213, USA}\\
\normalsize{$^{2}$Machine Learning Department, Carnegie Mellon University, Pittsburgh PA 15213, USA}\\
\\
\normalsize{$^\ast$To whom correspondence should be addressed; E-mail:  barati@cmu.edu.}
}

\date{}

\begin{document} 

\baselineskip24pt

\maketitle 


\begin{sciabstract}
Deep Reinforcement Learning (DRL) has emerged as a powerful control technique in robotic science. In contrast to control theory, DRL is more robust in the thorough exploration of the environment. This capability of DRL generates more human-like behaviour and intelligence when applied to the robots. To explore this capability, we designed challenging manipulation tasks to observe robots strategy to handle complex scenarios. We observed that robots not only perform tasks successfully, but also transpire a creative and non intuitive solution. We also observed robot's persistence in tasks that are close to success and its striking ability in discerning to continue or give up. 
\end{sciabstract}

\section*{Summary}

Creative robot manipulation can be achieved with deep reinforcement learning thorough exhaustive exploration of the environment.


\section*{Introduction}

With the emergence of novel artificial intelligence (AI) algorithms, and more specifically those in deep reinforcement learning (DRL), there will be a paradigm shift in robot perception and intelligence. In the future, AI enabled robots will be able to learn more quickly, generate complex behavioral strategies, and even achieve human level performance \cite{Mnih2015,Silver1140,OpenAI_dota,alphastarblog}. For many years, robots have been using control theory successfully to perform tasks. However, control theory has certain limitations. For example, robots depend heavily on human understanding of the task, the robot itself, and the environment. While there are promising results showing adaptability improvements \cite{Adaptive_Control}, robots still assume perfect knowledge of the system's description and the environment. On the other hand, DRL techniques, combined with proper training, enable robots to comprehensively explore the environment and learn appropriate solutions \cite{kober2013reinforcement}. The process of exploring the environment is a systematic advantage that we seek to exploit in order to generate creative solutions. To this end, we define creativity as the ability to autonomously find novel solutions to overcome obstacles in order to achieve a given goal.


\begin{figure}[h]
  \centering
 {\includegraphics[width=0.3\textwidth]{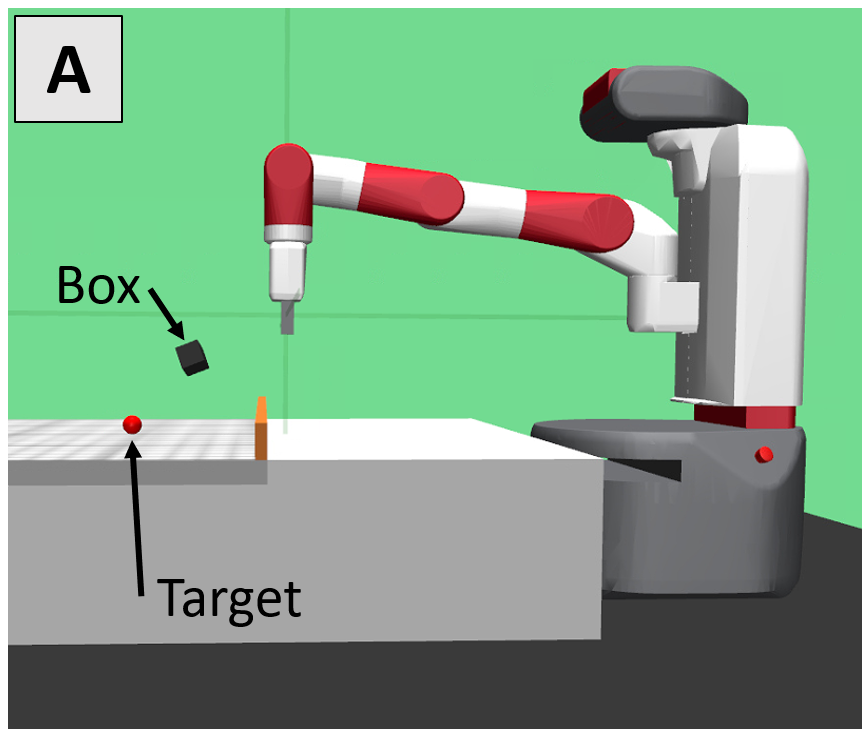}} 
 {\includegraphics[width=0.3\textwidth]{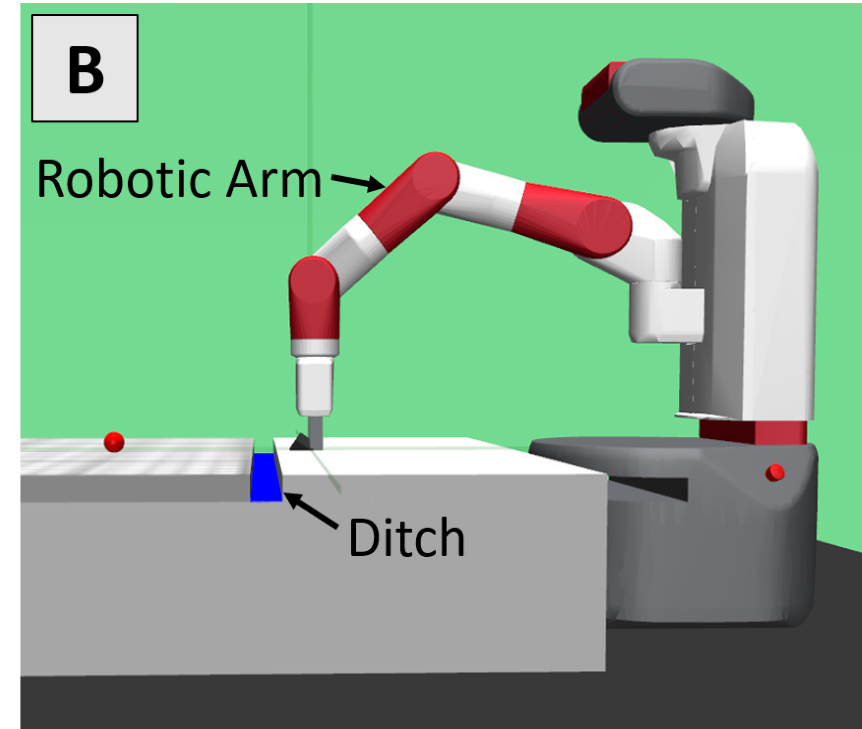}}
 {\includegraphics[width=0.3\textwidth]{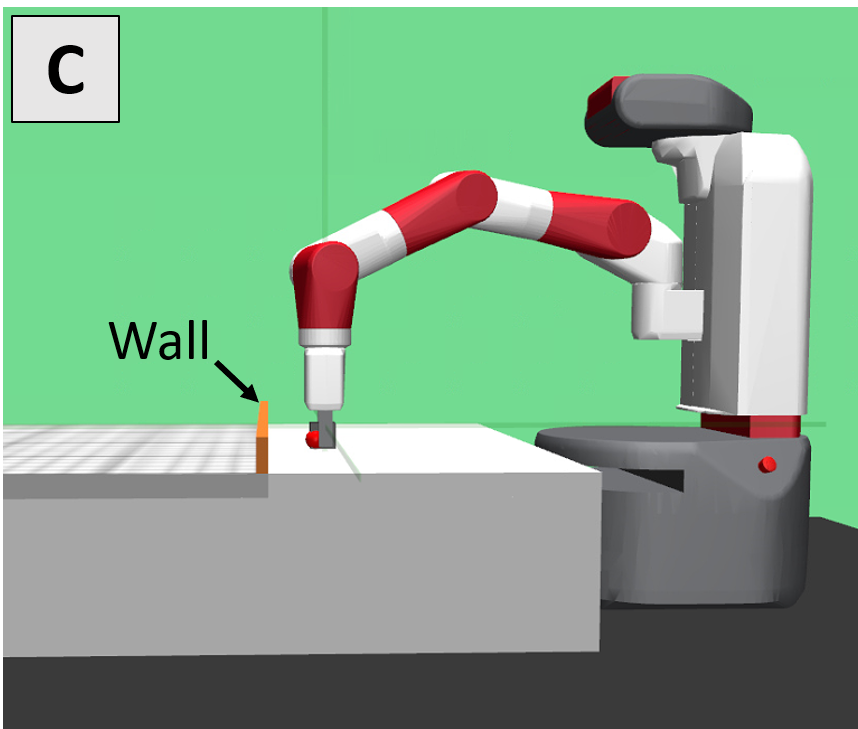}}
  \caption{Representation of the \texttt{Wall}, \texttt{Ditch}, and \texttt{Wall-TargetNear} environments. On the left the robot can be seen throwing the box to the target location. In the middle, the robot is about to make the box bounce over the ditch to the target by punching. On the right, the robot simply moves the box to the desired location, within the length of its arm. These figures are created using the MuJoCo physics engine \cite{todorov2012mujoco}.}
  \label{fig:experiment_environments}
\end{figure}

In this paper, we seek to show that a robot can find striking creative solutions during training. To demonstrate such capabilities, we performed 3 types of experiments using DRL applied to an industrial manipulator in a simulated experiment (Fig. \ref{fig:experiment_environments}). All three experiments require the robot to move a box to a the target location within the same table. We named the experiments \texttt{Ditch}, \texttt{Wall}, and \texttt{Wall-TargetNear} (See Supporting Information (SI)). We have specifically designed these experiments with different spatial constraints and challenges for the robot to test its problem solving strategy. To maximize the challenge, the target location is placed beyond the physical reach of the robot at all times, except for \texttt{Wall-TargetNear}, in which the target location is on an expanded range that includes closer locations. We used negative sparse rewards as an incentive for quick exploration. Through these experiments, we observed that the robot makes surprising decisions and finds creative solutions. 


\section*{Results}

\begin{figure}[h]
  \centering
 {\includegraphics[width=0.13\textwidth]{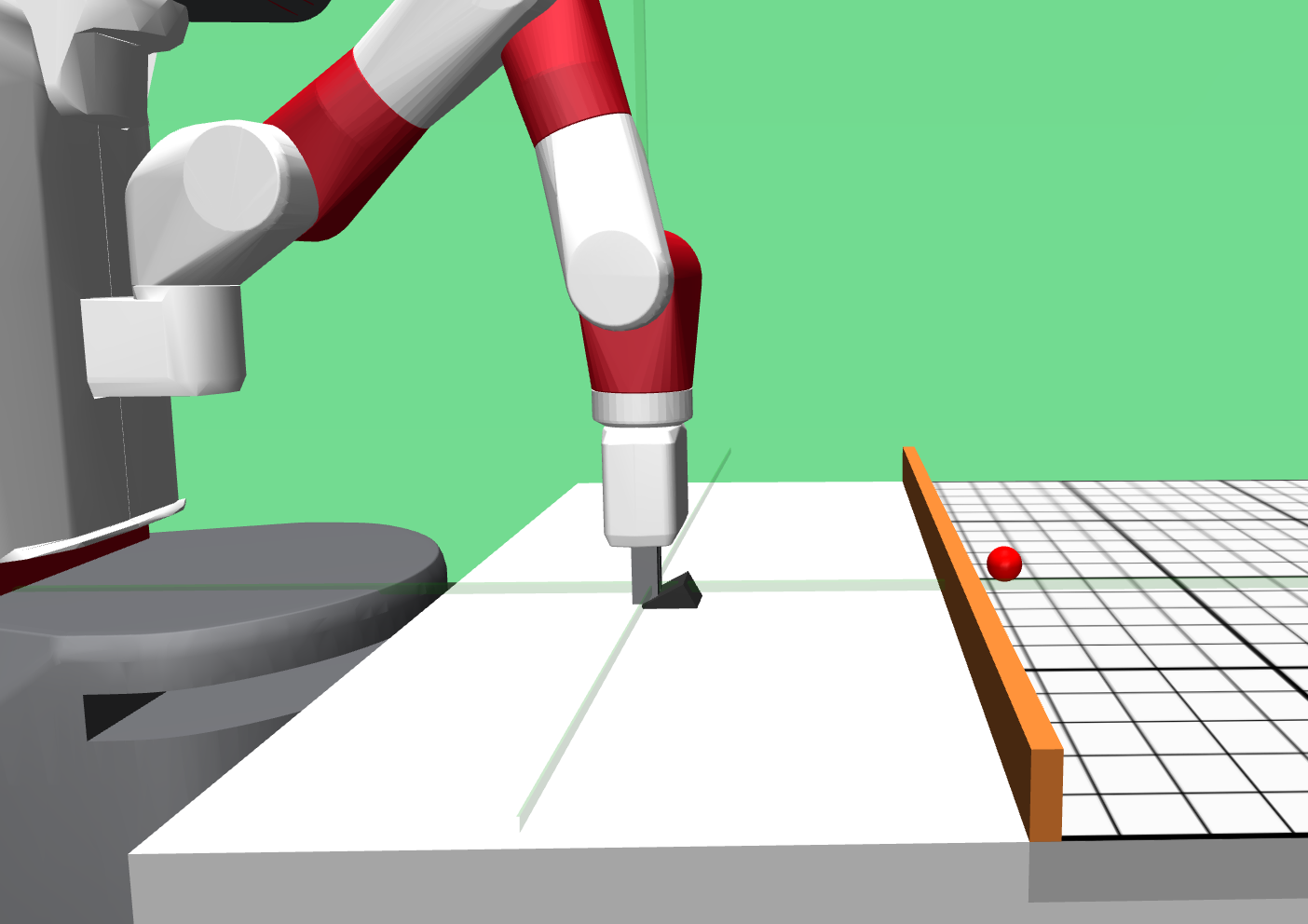}}
 {\includegraphics[width=0.13\textwidth]{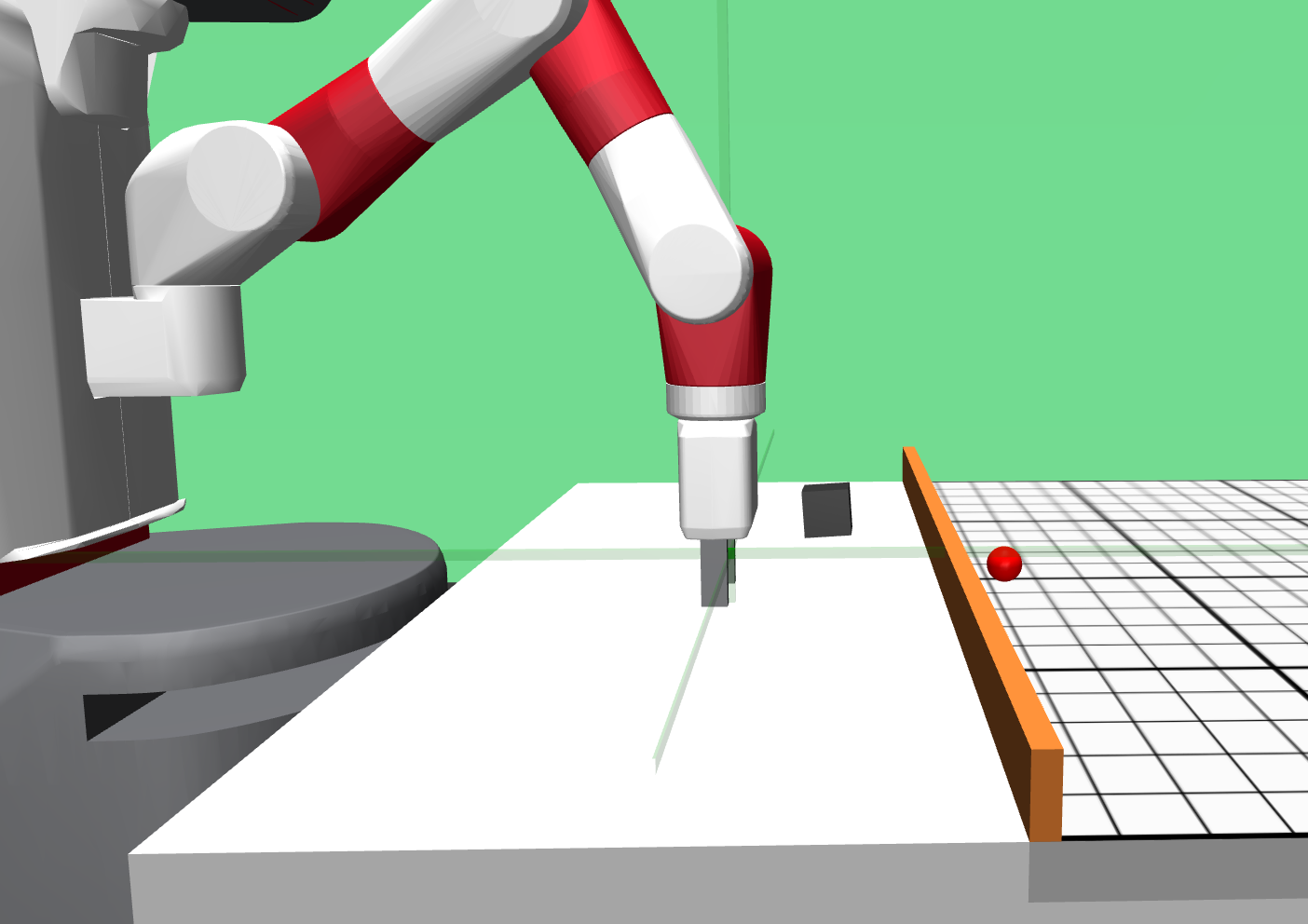}}
 {\includegraphics[width=0.13\textwidth]{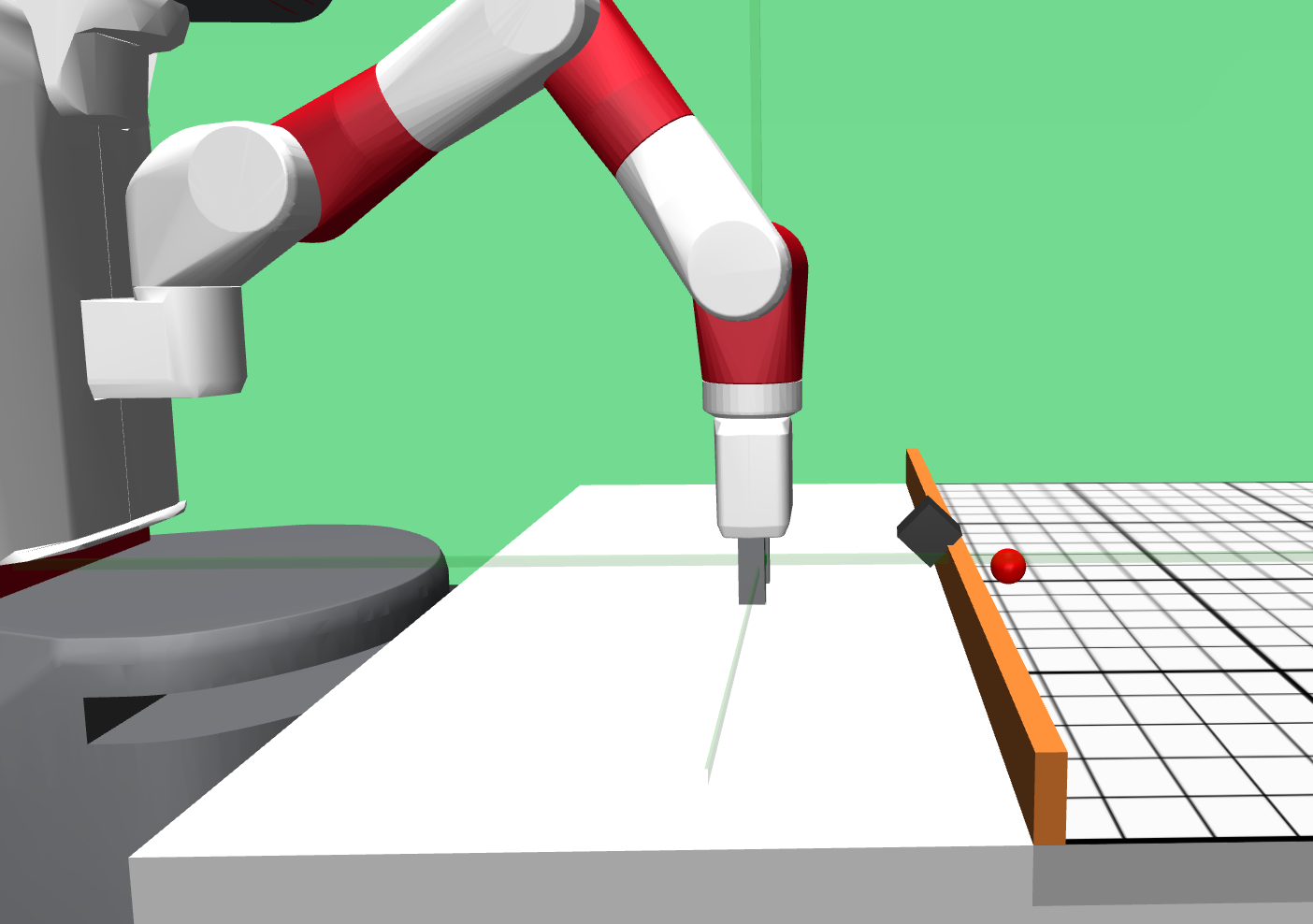}}
 {\includegraphics[width=0.13\textwidth]{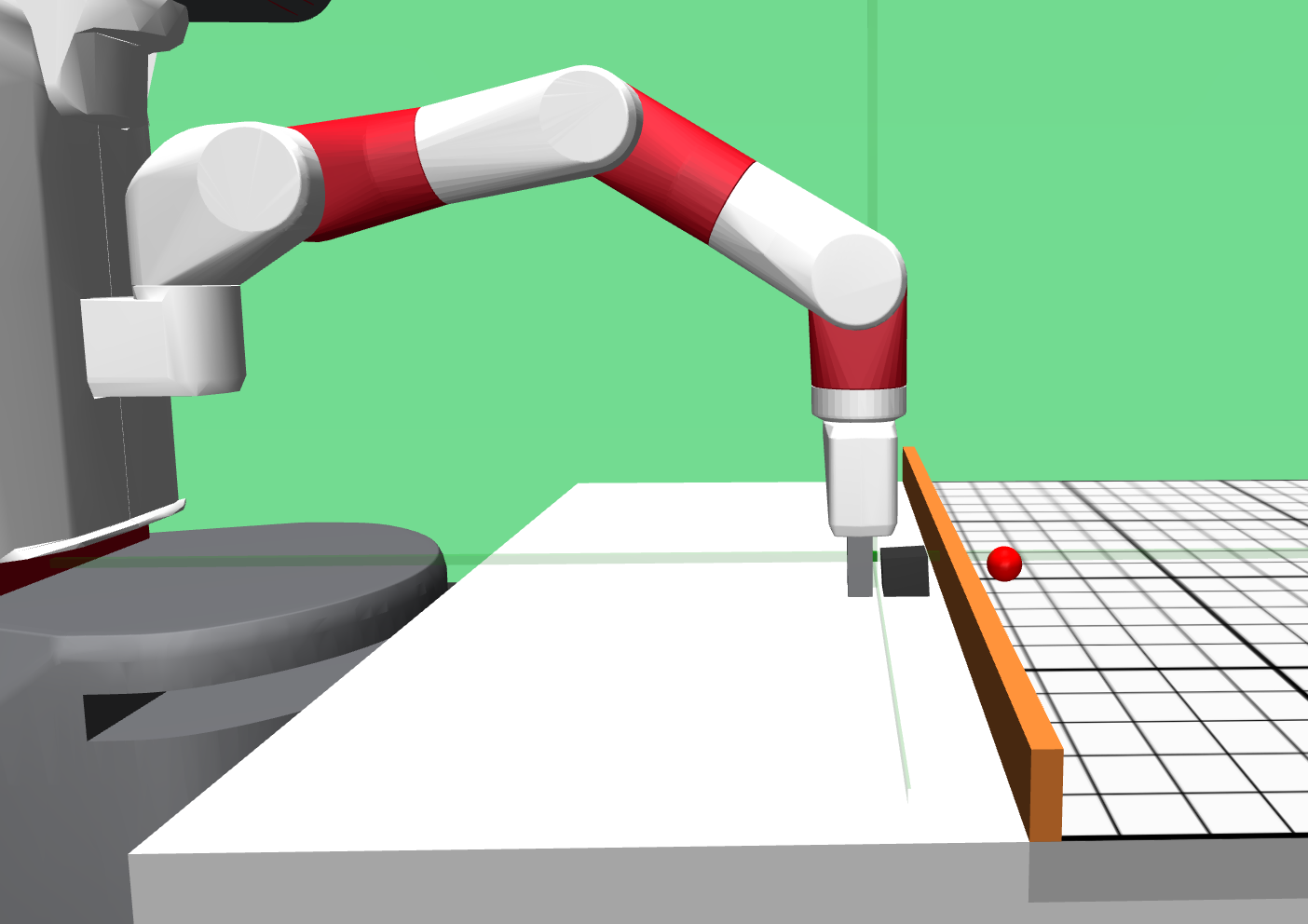}}
 {\includegraphics[width=0.13\textwidth]{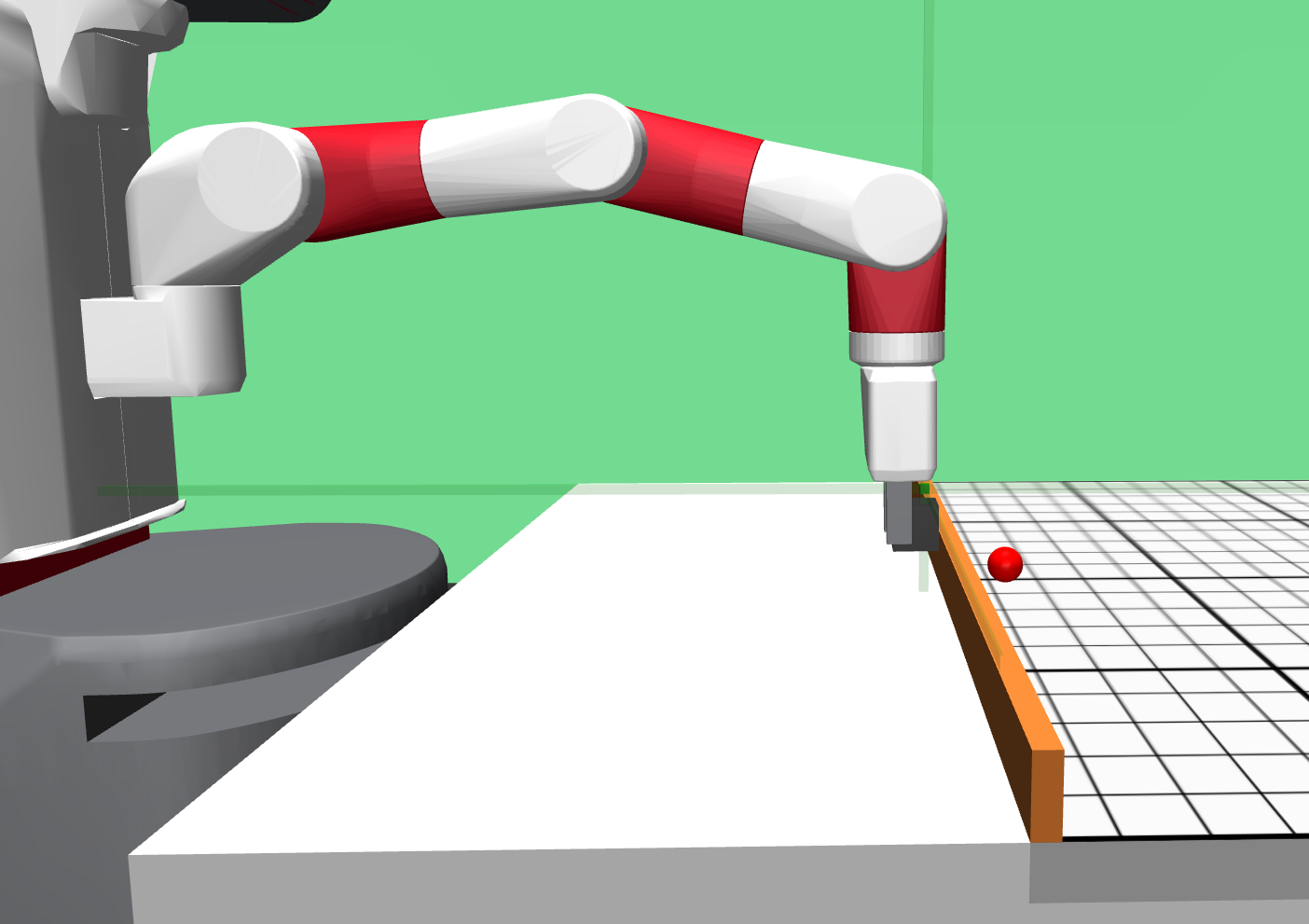}}
 {\includegraphics[width=0.13\textwidth]{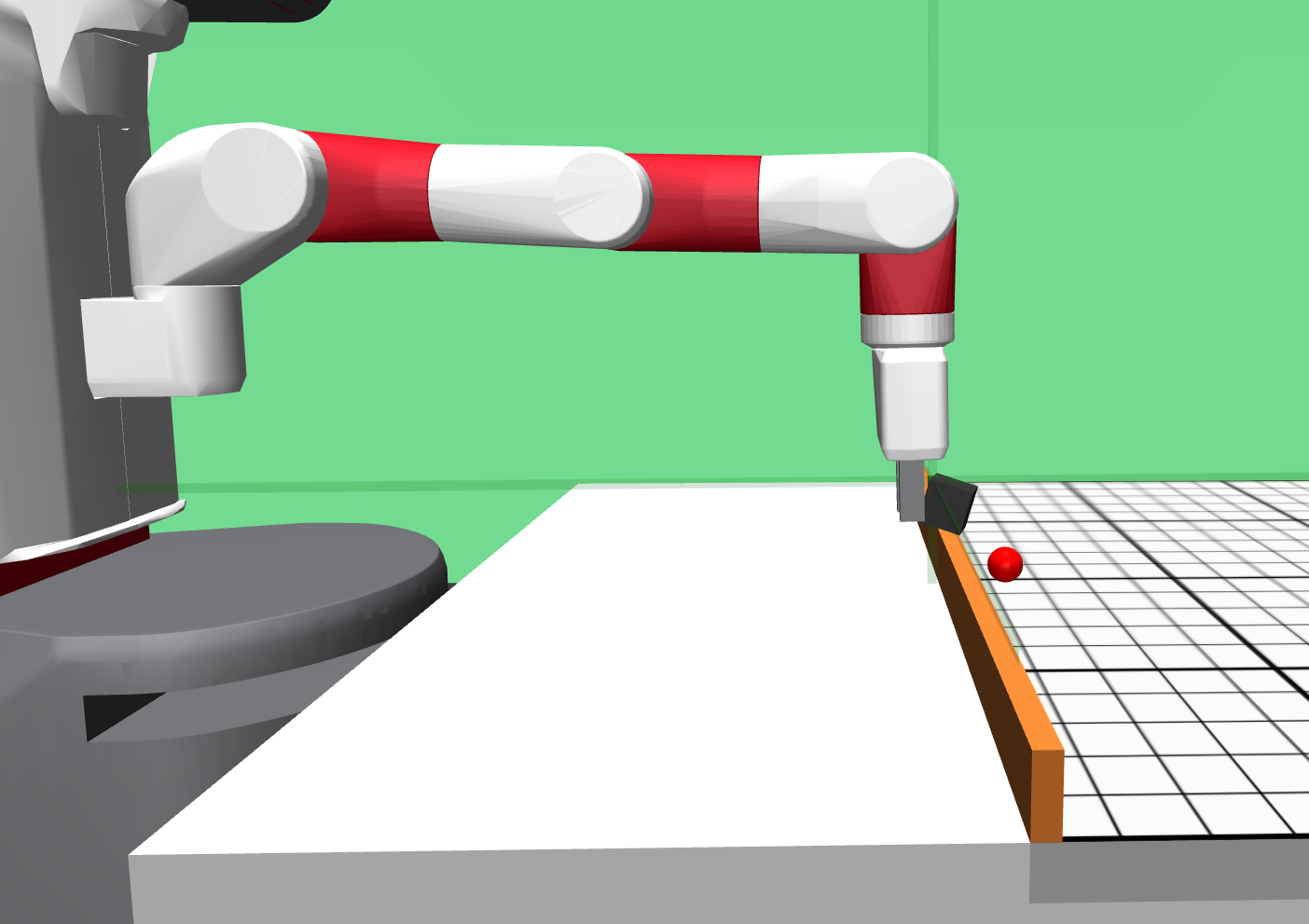}}
 {\includegraphics[width=0.13\textwidth]{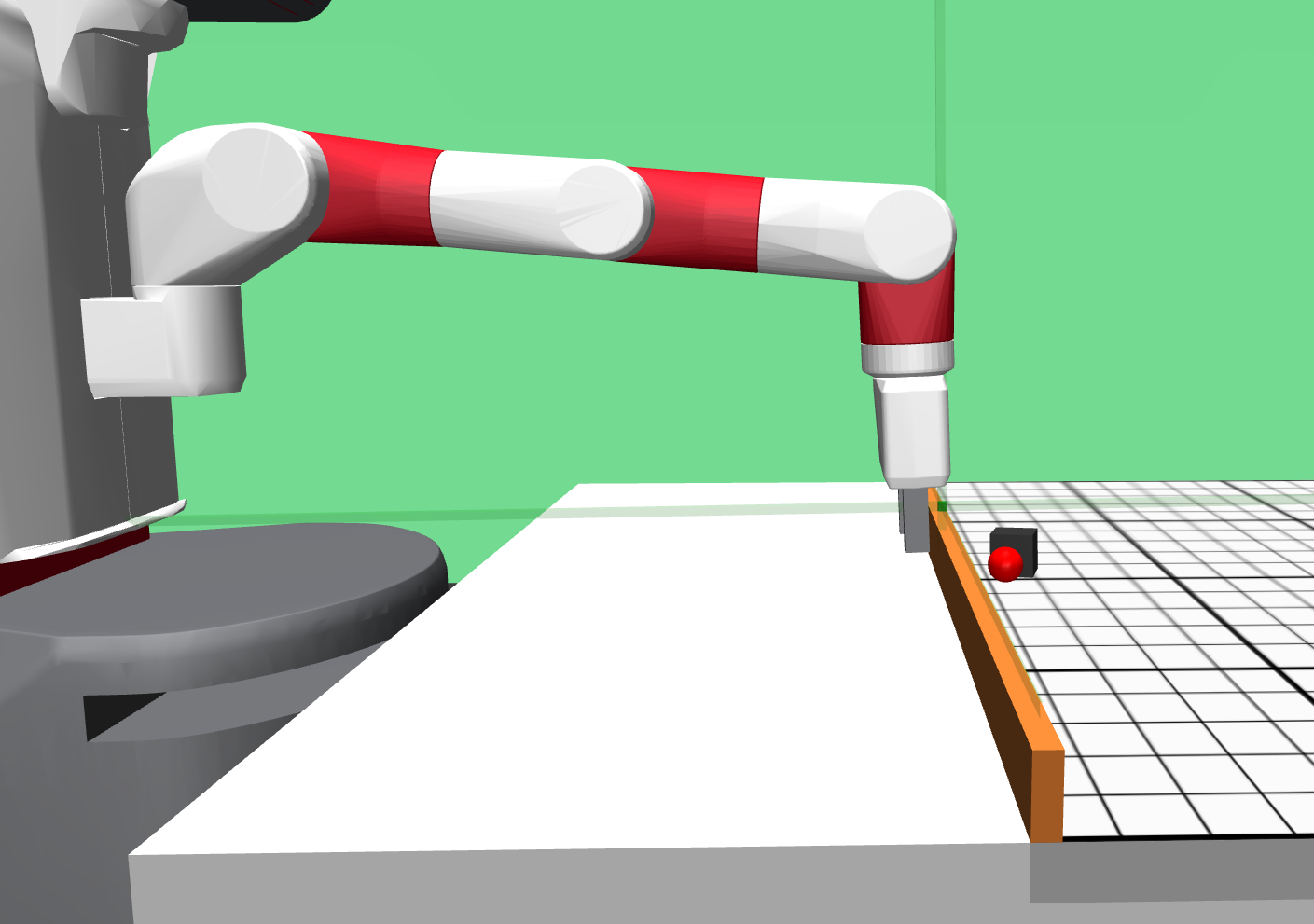}}
 {\includegraphics[width=0.13\textwidth]{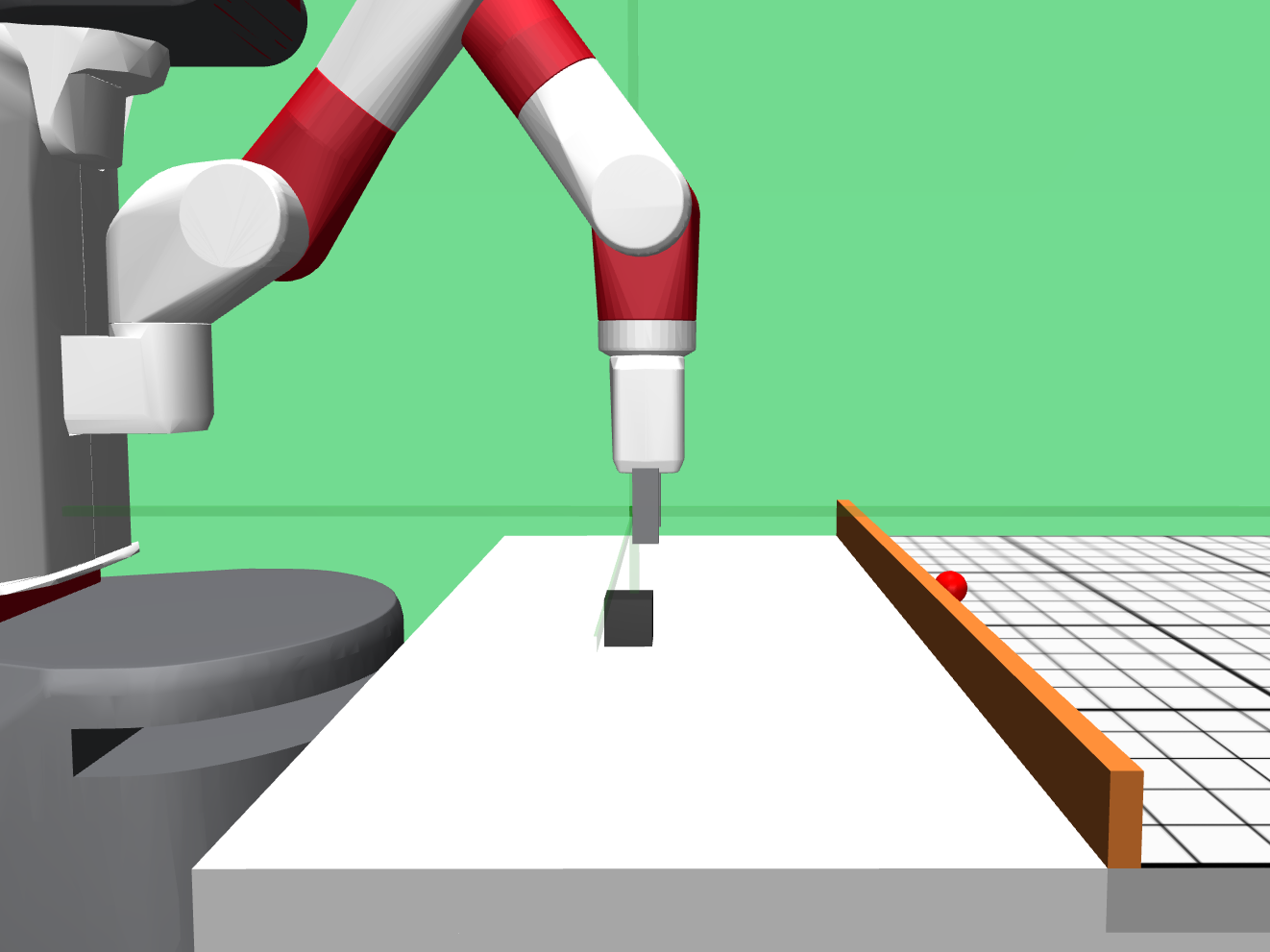}}
 {\includegraphics[width=0.13\textwidth]{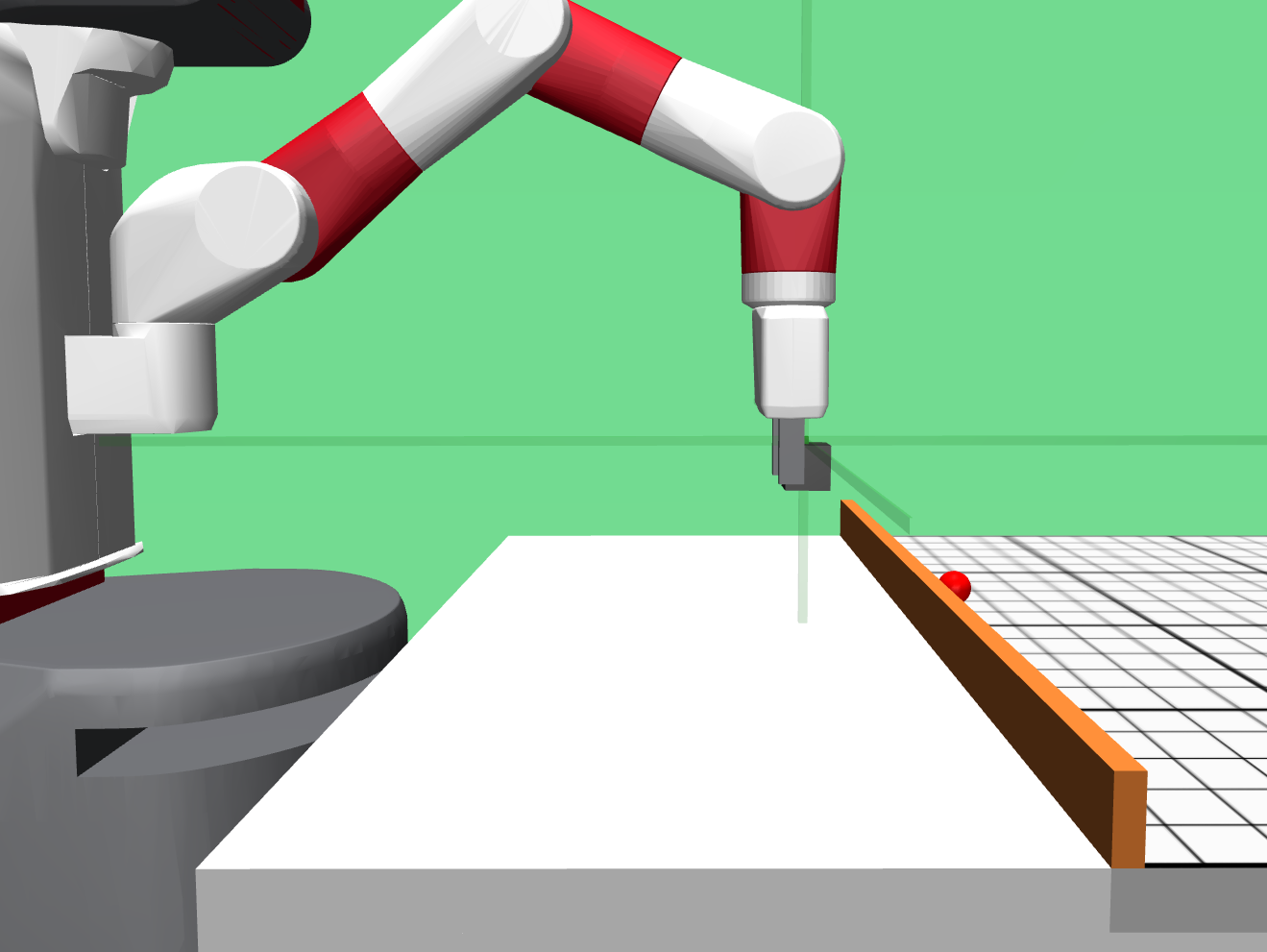}}
 {\includegraphics[width=0.13\textwidth]{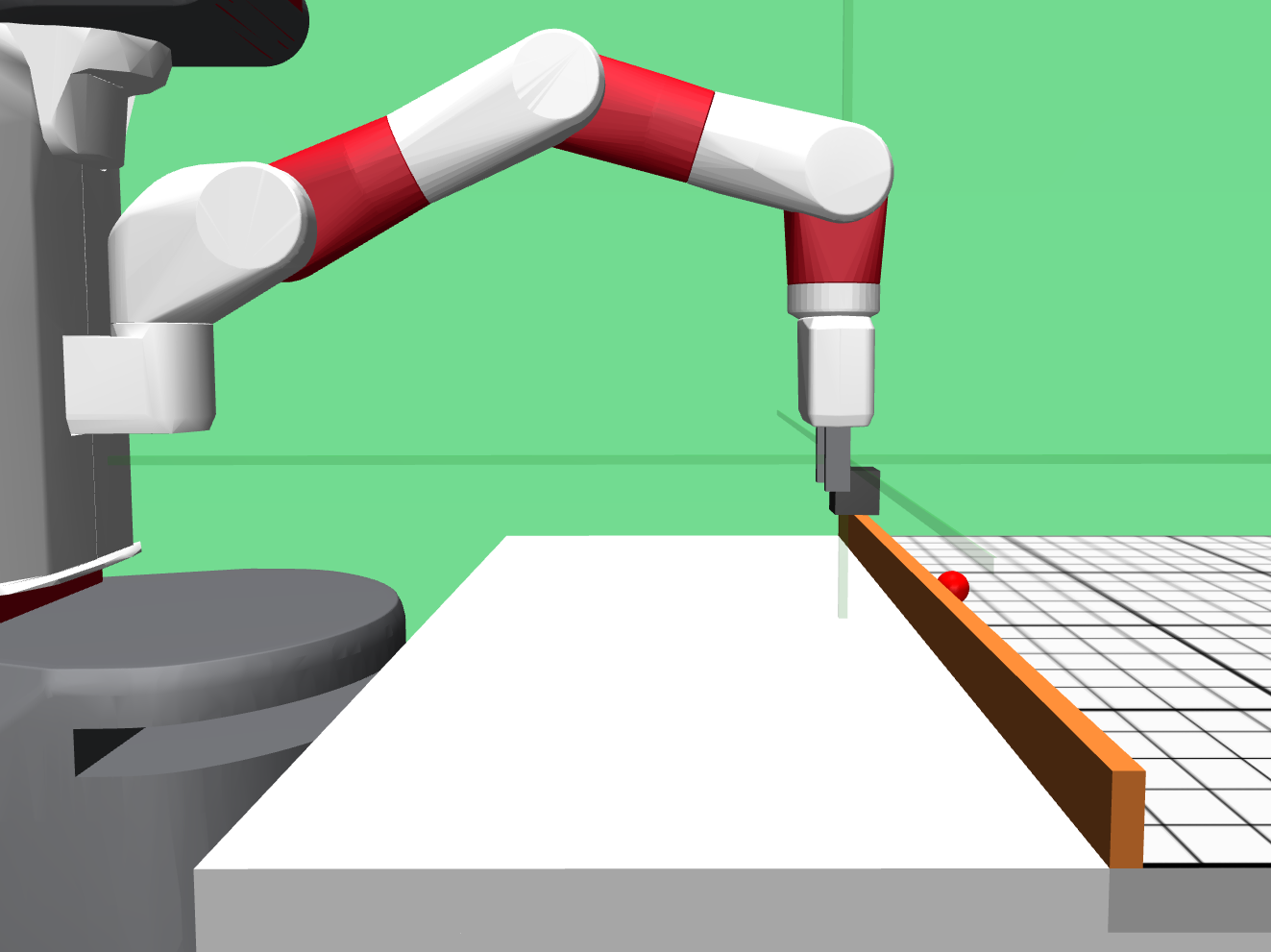}}
 {\includegraphics[width=0.13\textwidth]{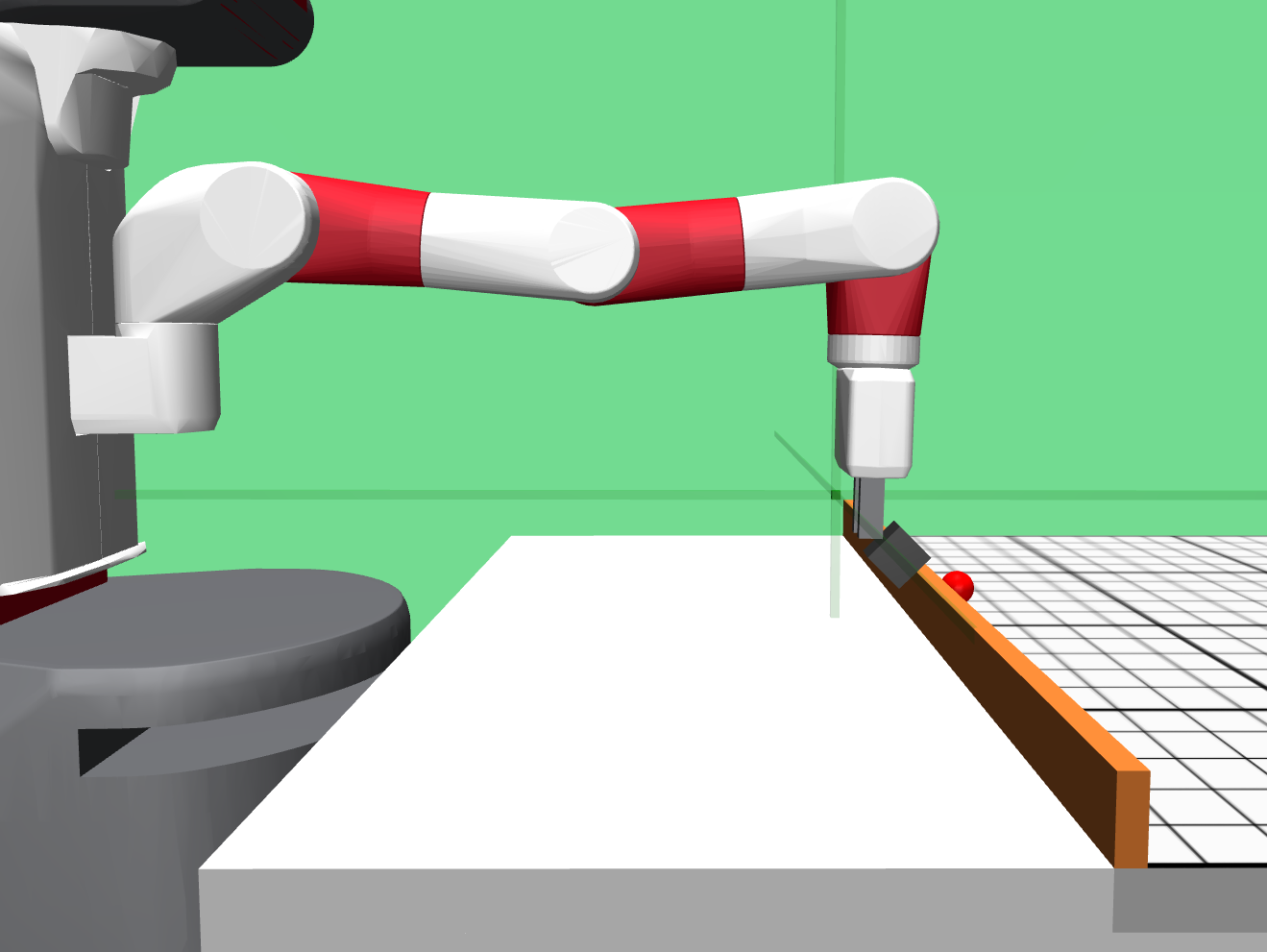}}
 {\includegraphics[width=0.13\textwidth]{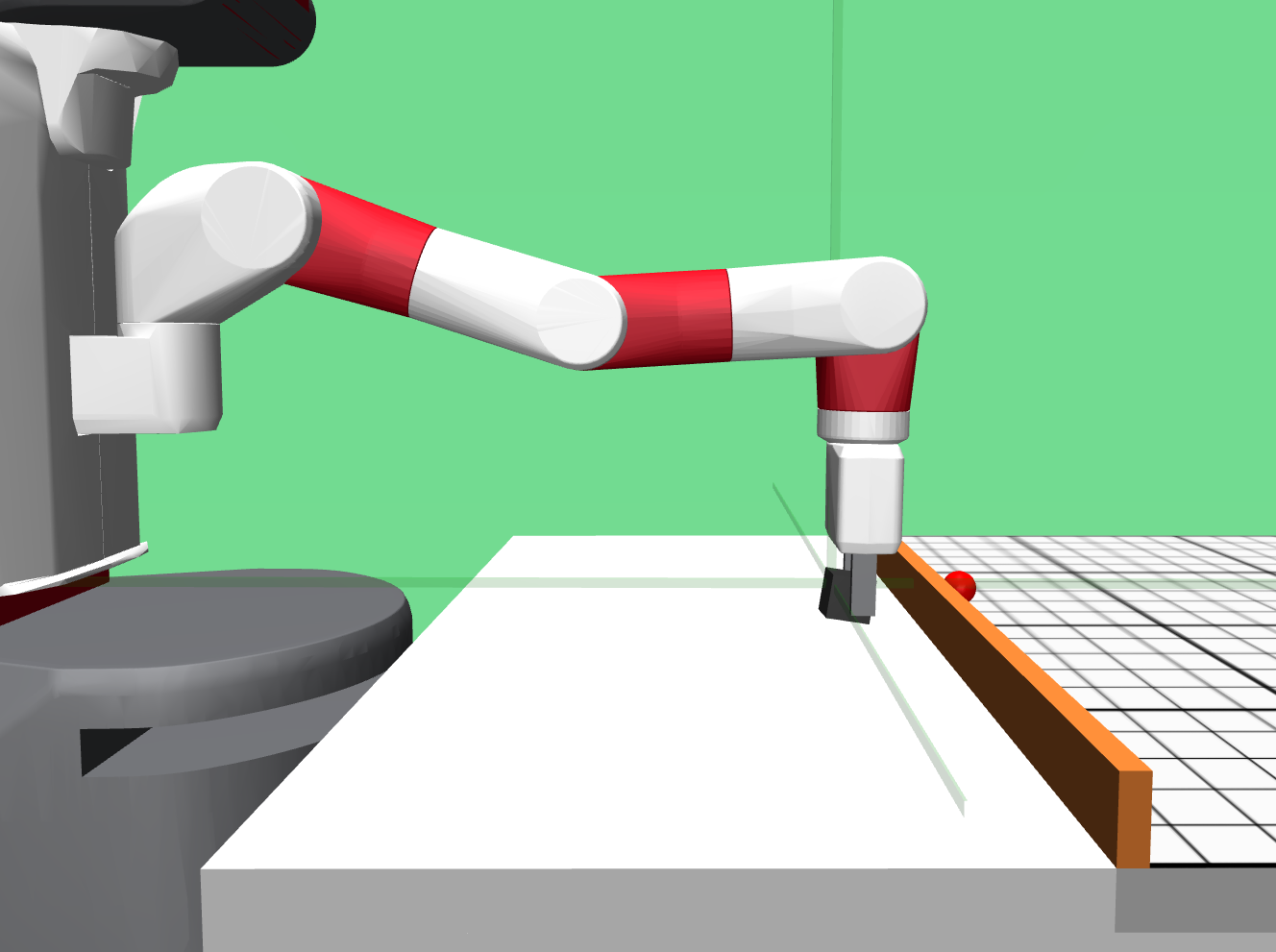}}
 {\includegraphics[width=0.13\textwidth]{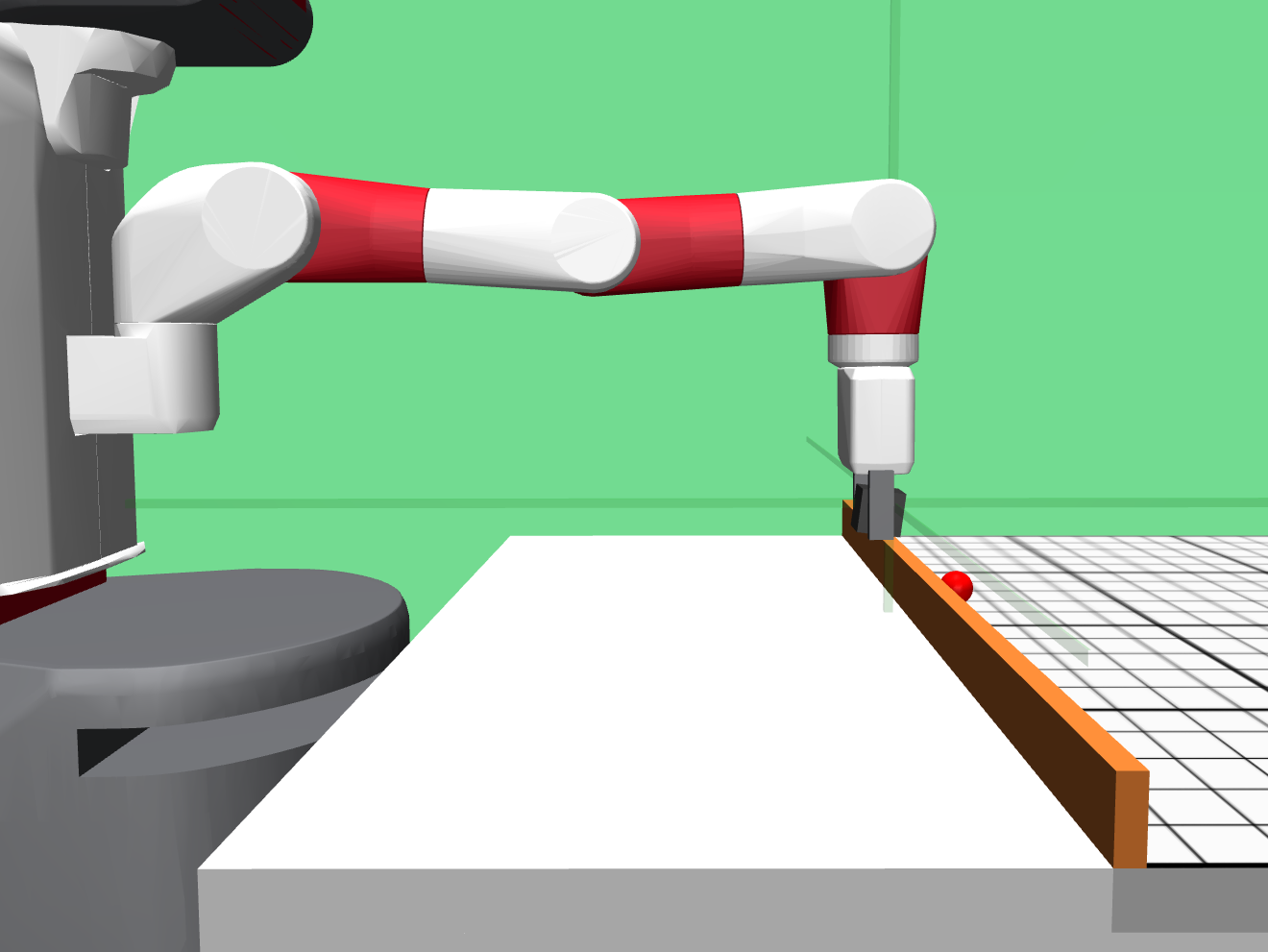}}
 {\includegraphics[width=0.13\textwidth]{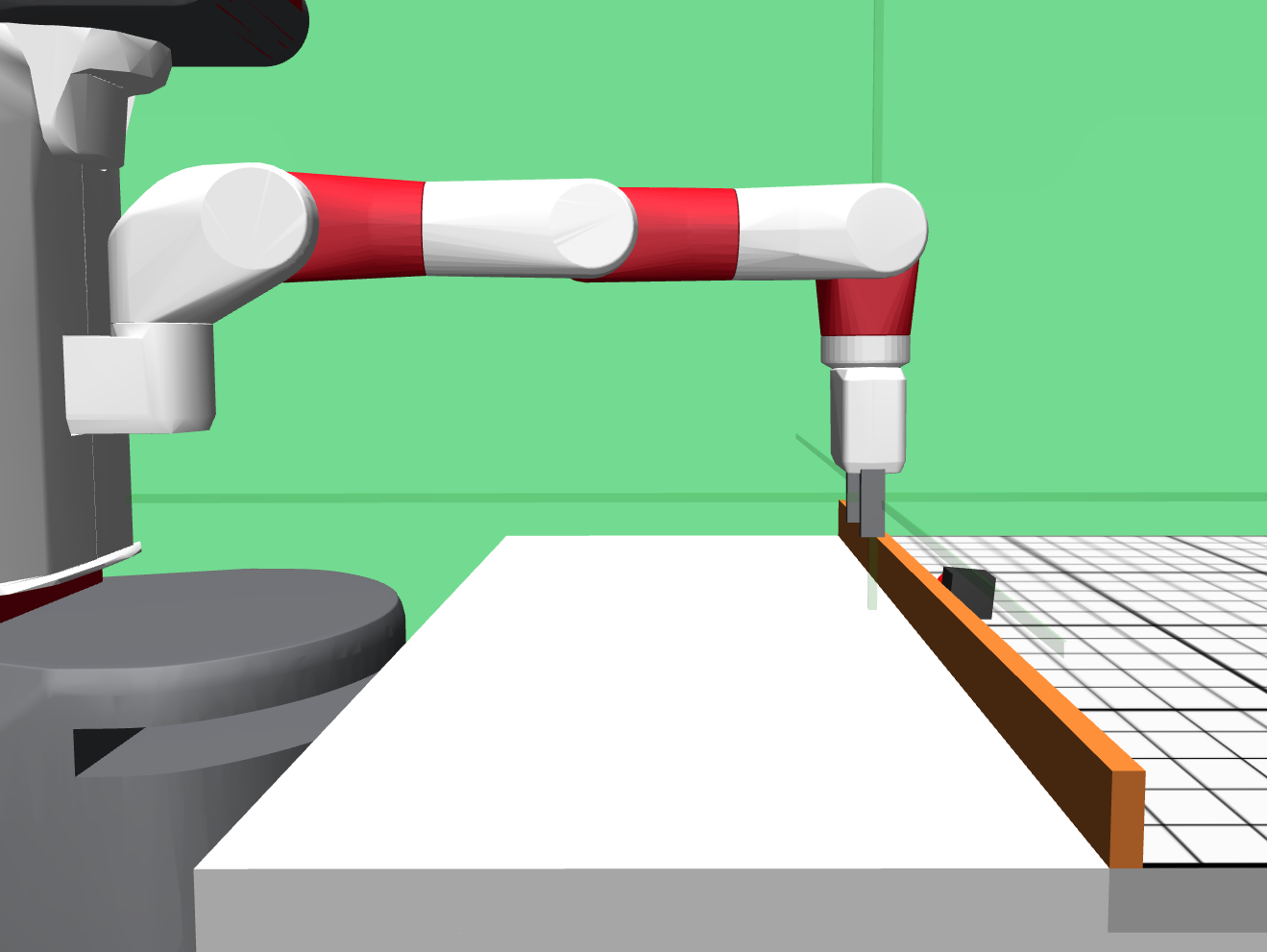}}
  \caption{The first row (a) shows a combined action of bouncing the box off the ground and then throwing over the wall. The second row (b) shows two subsequent throws after the first one misses.}
  \label{fig:episode_pointgrab}
\end{figure}

We observed that the robot became creative and found a solution that we did not expect. For all three experiments, due to the wall and ditch constraints, the robot needed to devise a solution to placing the box at the target location other than sliding or rolling it. Obviously, sliding or rolling would cause the box to either hit the wall or fall into the ditch. Our intuition was that the robot would learn to throw in order to get the box to the target location. However, the robot discovered that the box material is flexible and utilized that property to come up with a creative solution that is quicker to learn than throwing (throwing requires more dexterity with the gripper). \textit{Surprisingly, the robot learns to punch the flexible box against the table and, upon release, use the stored elastic energy to catapult the box over the constraint and into the target area.} (Figs. \ref{fig:experiment_environments}b and \ref{fig:episode_pointgrab}a and see supporting movies). The intriguing fact is that the physical properties of the box, such as the elasticity, weight, or friction between surfaces; are not part of the observation space. We believe discovering this hidden solution is valuable, creative, and well conceived (see supporting videos). For instance, the robot learned where and how much to press the box to store sufficient elastic energy to land it into the target. 

\begin{figure}[ht]
\centering
\includegraphics[width=0.98\linewidth]{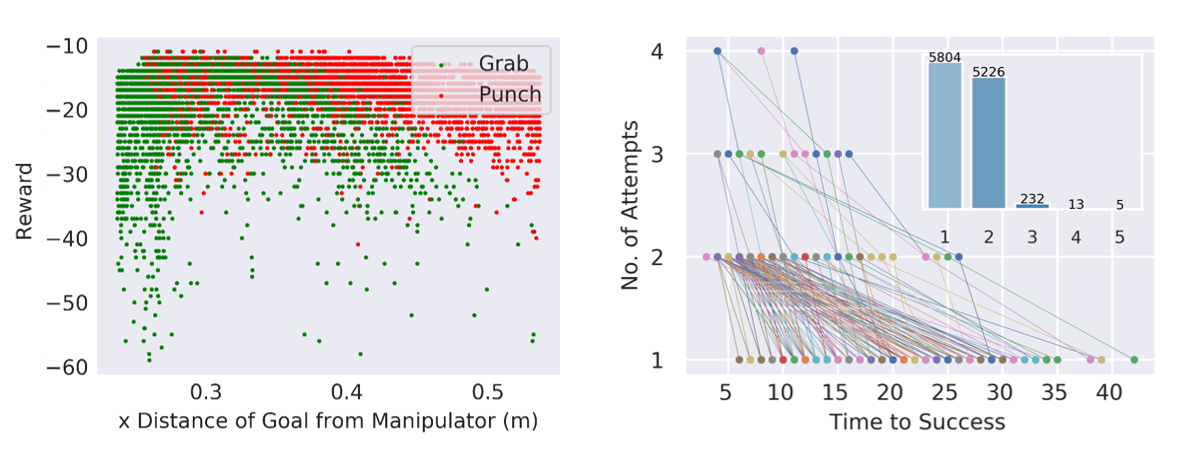}

\caption{(a) Sample policy learned for the \texttt{Wall} environment, filtered for successful episodes, and shown as a function of the distance of target to goal (x-axis) and the rewards obtained (y-axis). It shows the distribution of \textit{punch} vs. \textit{grab} as a function of distance. (b) Diagram of the number of attempts per episode vs. time steps left in an episode (extracted from policy in (a)) shows majority of episodes are successful within the first two attempts, although more attempts are possible if there are sufficient time steps left.}
\label{fig:success_strategy}
\end{figure}

As the robot gains more experience the actions become more complex. For instance, even as the robot already knows how to punch, it learns how to use its gripper to grab and eventually to throw the box. Evidence that it continues to explore the space even as a solution has already been found. Throwing often (but not always) generates more rewards than punching. This is very noticeable in the \texttt{Wall} experiment, in which the constraint is on the way of the trajectory for nearby punches (Fig. \ref{fig:success_strategy}a). The further away the target is placed, the robot is more likely to choose to punch. For other regions in which both actions are as likely to succeed, the robot is constantly and persistently looking for whichever action yields the highest reward, and in our case a hard preference based on distance was not clearly defined. Precision is also a factor that improved with training. The robot over time learns to accomplish the tasks using less individual actions (getting closer to a reward of $-10$), as opposed to multiple attempts (closer to a reward of $-60$) (Fig. \ref{fig:success_strategy}a). Therefore, the amount of reward that is accumulated on each episode is not only a metric of success, but also a metric of how quickly the robot can complete the task. In the event that the first attempt was unsuccessful (Fig. \ref{fig:success_strategy}b), and the box is still within reach of the gripper, the robot will pursue multiple attempts until the time runs out or it is successful. This behavior is a byproduct of using negative sparse rewards.

\begin{figure}[ht]
\centering
\includegraphics[width=0.98\linewidth]{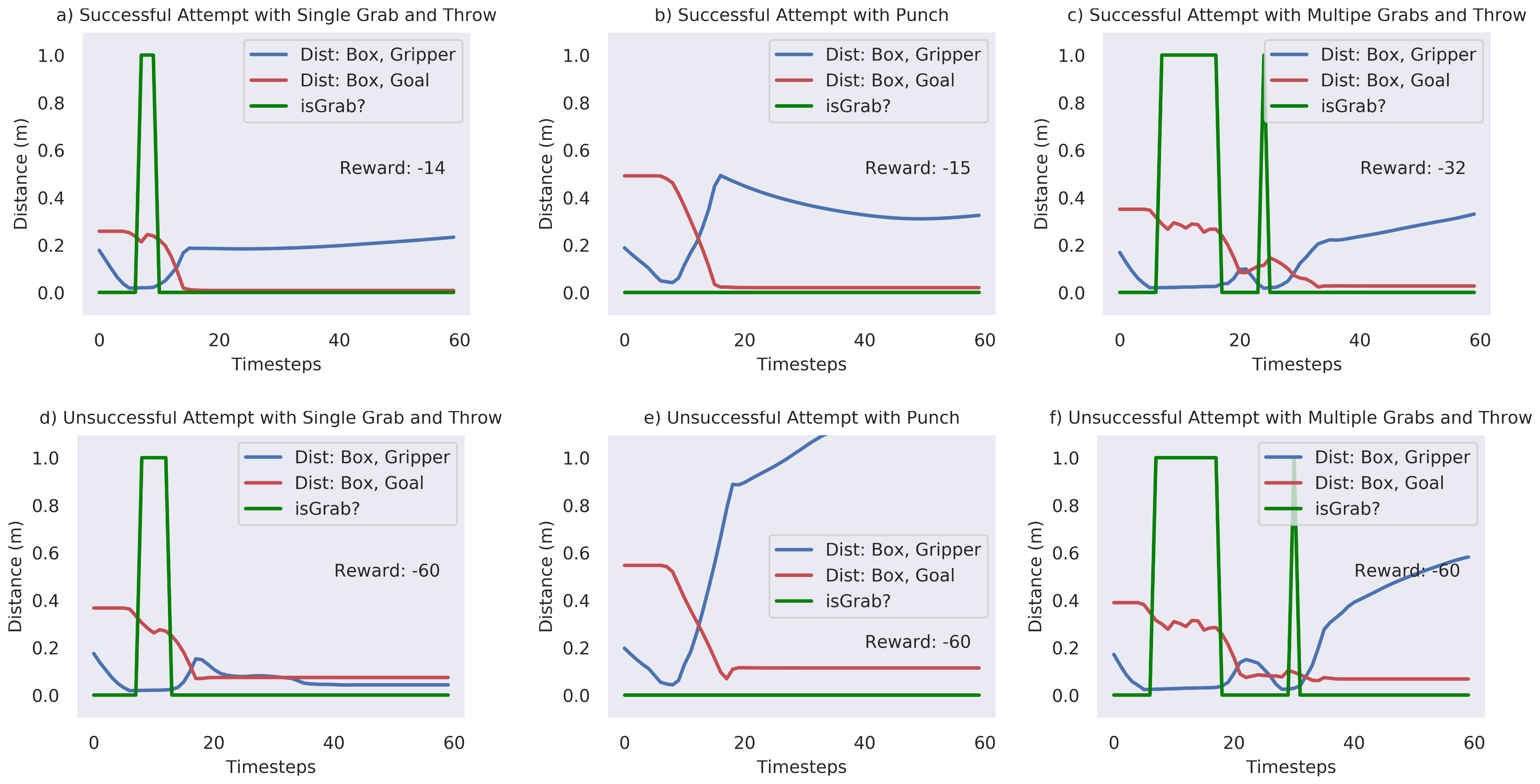}
\caption{Analysis of single episodes for the \texttt{Wall} experiment when the robot (a) grabs and throws the box right on target, (b) punches the box on target, (c) executes two grabs and throws on target, (d) throws the box and misses, (e) punches the box and misses, and (f) grabs a couple of times before throwing but misses. The tolerance for being on target is $0.05m$.}
\label{fig:combined}
\end{figure}

For the robot, a creative solution is not sufficient. Mastery of each one of the actions is paramount to achieve higher rewards. Deep inside each episode we can see that the amount of time each action takes has to be small in order to accumulate higher rewards. All combined, the reward is unlikely to be ever higher than $-10$. The worst reward, $-60$, comes from the maximum number of time steps that we allow per episode, and each episode lasts $2.76$ seconds. Similarly, for the best case, the gripper takes at least about 6 time steps, or $0.276$ seconds, to move from its starting position to the box, leaving an additional 4 or 5 steps, merely a fraction of a second, to throw or punch the box over the constraint (Fig. \ref{fig:combined}), and have it land (and stay) at the right location. Some high reward ideal scenarios are shown on Fig. \ref{fig:combined}a and Fig. \ref{fig:combined}b. Conversely, the scenarios on Fig. \ref{fig:combined}d and Fig. \ref{fig:combined}e are likely to be the worst, since a single (hasty) action that lands the box beyond the constraint cannot be corrected (the box sits beyond the reach of the robot). When the action is unlikely to be successful, the best outcome comes from performing multiple successive actions to re-position the box before sending it over the constraint. This improves the chances of success (and collect some reward), even if it is at the expense of time.  

\begin{figure}[h]
\centering
\includegraphics[width=0.98\linewidth]{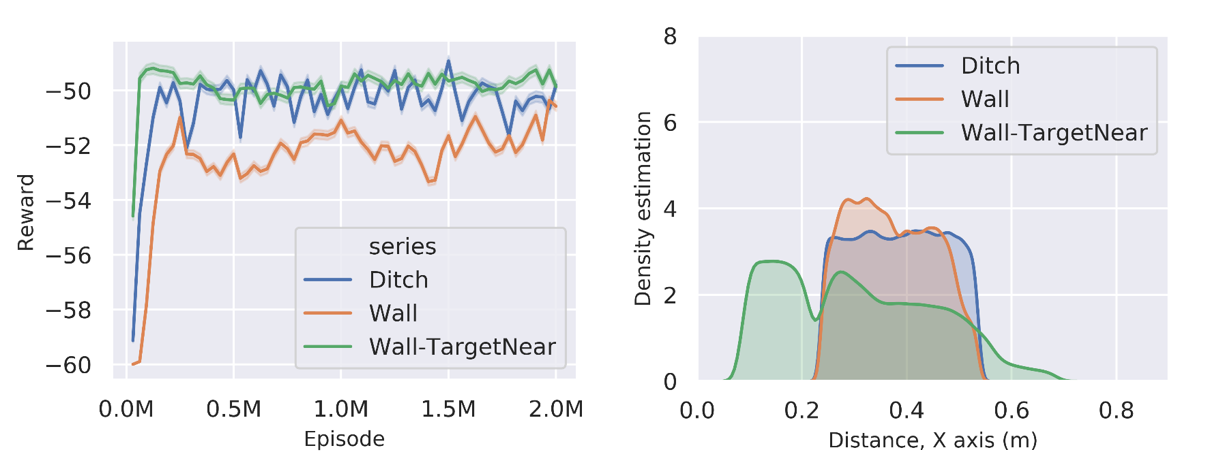}
\caption{(a) The average episodic rewards for each one of the three experiments during training (for both successful and unsuccessful episodes). (b) Density estimate plots along the X axis show the effect of constraints in the learning and creative process. Given the uniform target sampling distribution, horizontal densities would be ideally expected. The robot on the \texttt{Wall-TargetNear} experiment has become more successful at clear proximity. The robot on the \texttt{Wall} experiment performed better close to the barrier. The robot on the \texttt{Ditch} experiment performed close to uniform throughout the range.}
\label{fig:experiments_001}
\end{figure}

The robot displayed a similar level of creativity for the \texttt{Ditch}, \texttt{Wall}, and \texttt{Wall-TargetNear} experiments, as measured by learning to grab and punch the box. Adding environmental constraints to the process of reaching the goal affected only marginally the time needed to learn how to punch or grab the box (Fig. \ref{fig:experiments_001}a). However when faced with different environmental constraints, the robot modifies its policy to maximize its rewards (Fig. \ref{fig:experiments_001}b). For instance, for the \texttt{Wall-TargetNear} experiment, the robot was consistently successful at the closer range, but less so at distances farther away beyond the wall. On the other hand, in the \texttt{Ditch} experiment, the robot was almost equally successful across the entire range ($25cm$ to $55 cm$ range along the long side of the table), which we interpret as the target having a uniform difficulty level. In all cases we utilized uniform sampling of the target space.

\section*{Discussion}


 We have shown that deep reinforcement learning can be used as a technique to explore the environment and construct creative solutions for complex scenarios. We demonstrated that the robot is able to learn and use physical properties of environment, such as the elasticity, friction, and weight of the box for solving difficult manipulation tasks. These properties are not given explicitly as the states, however; robot was able to discover and use those to achieve its goal. Finding the punch strategy to throw the box to the target location is novel, creative and non intuitive to human. This makes us reflect about the importance of allowing robots to learn and be creative with as little guidance as possible. 
In addition, we observed robot persistence, another human like behaviour, in its confrontation with the failure. In many instances, the robot assess that it can be successful if it tries more. The time to success analysis in this paper shows how robot iteratively gets closer to success and is persistant.

In this study, we are limited by the quality of the simulated environment. For our experiments we also rely on an unchanging test framework in which, for instance, the target area remains the same to enable uniform transition sampling across the observation space and avoid catastrophic forgetfulness. These issues fall outside of the scope of this set of experiments.


\section*{Materials and Methods}

We conduct the experiments on a modified version of OpenAI's Fetch-Slide environments \cite{DBLP:journals/corr/abs-1802-09464}, powered by the MuJoCo physics engine \cite{todorov2012mujoco}. The setup involves a robotic manipulator trying to move a cubic object (box) to a particular location (target) (Fig. \ref{fig:experiment_environments}). The experiments involve adding different constraints to this environment, modifying the goal, and thereon exploring the robot's behavior. We want to see if the robot is able to adapt to the additional constraints, and whether it is creative enough to learn new policies by comprehensively exploring its action space. For this purpose we created 3 variations of two slightly different environments. For the first environment we created a \textit{ditch} to constrain the robot from rolling or pushing the box to the goal, or target location (the robot should learn to avoid dropping the box into the ditch). (Fig. \ref{fig:experiment_environments}) Second, we created a \textit{wall} to constrain the robot and increase the difficulty of the task (the wall is far enough that it cannot be reached by the robot, forcing it to find a way to pass the box over the wall). The environment with the wall constraint was modified further by creating additional distance variations (see SI).

For the architecture of DRL, we use Deep Deterministic Policy Gradients (DDPG) \cite{DBLP:journals/corr/LillicrapHPHETS15}  with Hindsight Experience Replay (HER) \cite{DBLP:journals/corr/AndrychowiczWRS17}. We chose DDPG with HER because our experiments require both a continuous action space and a continuous state space. DDPG uses an Actor-Critic network (see SI), in which the actor's policy is a Deterministic Policy Gradient network and the critic's policy is a Q-Value network. The state space is a vector of 25 parameters which include the position of the gripper [3 parameters], position of the box [3], rotation of the box [3], velocity of the box [3], rotational velocity of the box [3], relative position between the gripper and the box [3], state of the gripper [2], gripper positional velocity [3], and gripper velocity [2]. The \textit{goal} in the input space refers to the coordinates of the target [3]. The size of the action is four- comprising the 3D coordinates of the gripper and the gripper state (distance to the center). 

The actor network is comprised of 3 hidden layers of fully connected neurons (see SI). The input to the actor network is the combination of the state, the achieved goal, and the desired goal. The output is the action. The critic network is comprised of also 3 hidden layers of fully connected neurons. The input of the critic network is the state, the achieved goal, the desired goal, and the action (the output from the actor network). The output is the Q value evaluated for the current reward. The combination of DDPG and HER enables the robot to learn faster by learning from previous failure. We chose to use \textit{sparse} rewards for several reasons. First, the outcome of the robot's actions is either right or wrong (within a certain tolerance). Using \textit{dense} rewards could be misguiding the robot into performing incomplete actions to get at least some reward \cite{guo2017deep}. Second, we didn't want to invest time engineering rewards for specific use cases. By engineering a reward, we would be leading the robot into learning \textit{our} solution to the problem, hindering exploration \cite{DBLP:journals/corr/AndrychowiczWRS17}. Instead of leading the robot, by adopting a sparse reward strategy, we encourage an explore-first discover-later strategy for each episode. Lastly, providing negative rewards works as an incentive to achieve the goal as quickly as possible. Each episode lasts 60 time steps, and for each step the robot receives a $-1$ reward if the goal has not been achieved, or $0$ otherwise. As a consequence of this reward structure, the minimum cumulative reward per episode is $-60$, and the maximum theoretical cumulative reward is $0$.

We train on a single Intel Core i7-8700K CPU at 3.70GHz × 12 cores server with a GeForce GTX 1080 Ti/PCIe/SSE2 GPU and 16 Gb of RAM, running Ubuntu 16.04.  For all environments, we trained for $100 \times 10^{6}$ time steps. For policy analysis, approximately 500,000 test episode samples were collected from each of the experiments. Additionally, 20,000 extra test episode samples were collected with step by step information from the \texttt{Wall} experiment for episodic analysis.



\nocite{DBLP:journals/corr/MnihKSGAWR13,Experience_Replay,DBLP:journals/corr/LillicrapHPHETS15,degris2012off,DBLP:journals/corr/AndrychowiczWRS17,guo2017deep,Pathak2017CuriosityDrivenEB,Ng2000AlgorithmsFI,Ziebart2008MaximumEI,Abbeel2004ApprenticeshipLV}
\nocite{lecun2015deep}
\nocite{Goodfellow-et-al-2016}
\nocite{barto}
\nocite{8636075}
\nocite{silver2014deterministic}
\nocite{DBLP:journals/corr/SchulmanWDRK17}
\nocite{DBLP:journals/corr/WangBHMMKF16}
\nocite{DBLP:journals/corr/abs-1902-00528}
\nocite{schulman2015high}
\nocite{schaul2015prioritized}
\nocite{wawrzynski2013autonomous}
\nocite{ioffe2015batch}
\nocite{hasselt2010double}
\nocite{wawrzynski2009real}
\nocite{CNRL}
\nocite{guo2017deep}
\nocite{DBLP:journals/corr/abs-1802-09464}
\nocite{baselines}
\nocite{DBLP:journals/corr/abs-1808-00177}
\nocite{danica1,danica2}
\bibliography{references}

\bibliographystyle{Science}

\newpage

\section*{Supporting Information}

This section contains ancillary information beyond what is included in the main paper. 

\subsection*{Environment}

\begin{figure}[h]
\centering
\includegraphics[width=0.98\linewidth]{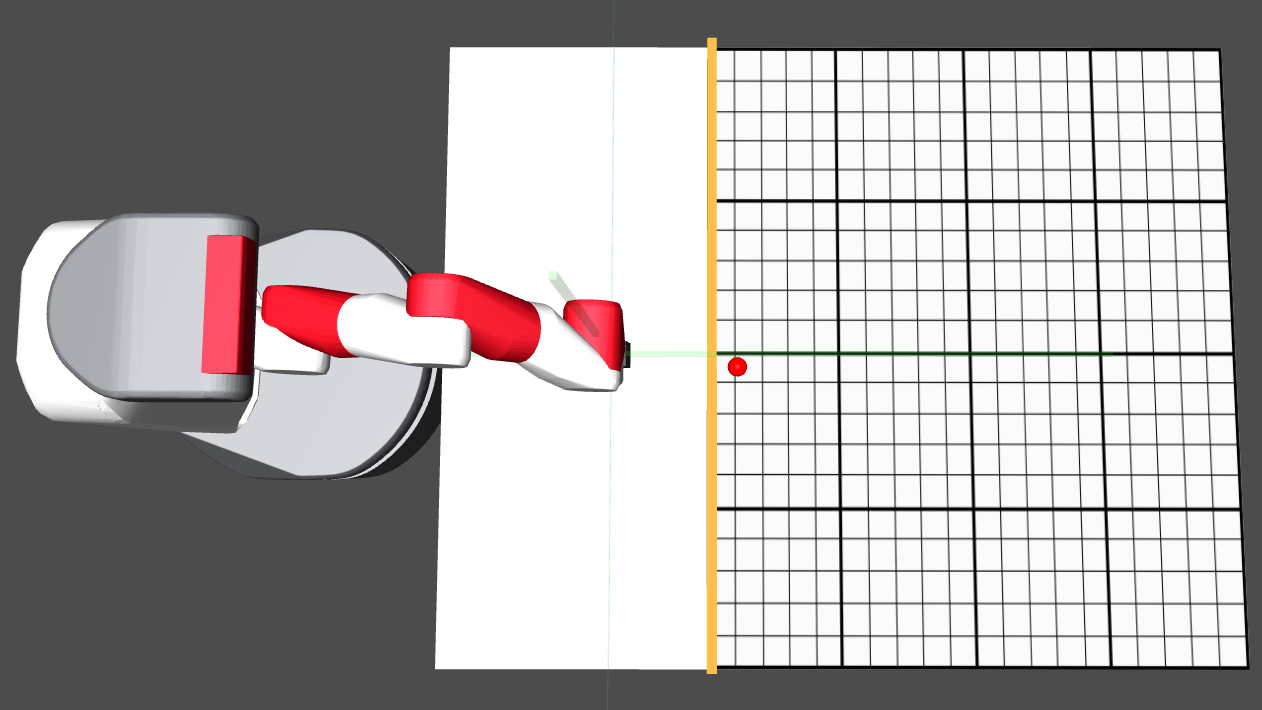}
\caption{Top view of environment used for \texttt{Wall} and \texttt{Wall-TargetNear} experiments. The initial position of the robot's gripper is the coordinate origin, while the direction left and right is along the 'x' axis, and the direction up and down is along the 'y' axis. Each one of the grid squares has an area of approximately $2.75 \times 3.25 cm^{2}$}
\label{fig:environment_topview}
\end{figure}

The robot sits approximately at the middle of the table on the 'y' direction (Fig. \ref{fig:environment_topview}) and cannot reach farther than the distance to the wall, shown in yellow (or the ditch, not shown). The range for the \texttt{Wall} and \texttt{Ditch} experiments is contained within the first two gridded regions from the left, and the middle two regions in the vertical direction. The entire target area is approximately a 4 region area. The area for the \texttt{Wall-TargetNear} experiment, on the other hand, is centered at the same point but the coverage has been expanded for about an entire region all around, including the area without a grid.

\subsection*{Configuration}

We use a standard DDPG+HER setup. Some of the most important hyperparameters are described below:
\begin{itemize}
\item Buffer size: number of transitions $\left(s_{t},a_{t},g_{t},r_{t},s_{t+1}\right)$ that are stored in the replay buffer.
\item Neurons per hidden layer: a definition of how many neurons will be available per hidden layer in the actor-critic network.
\item Number of layers: the number of hidden layers in the actor-critic network 
\item Network class: a definition of a neural network passed down as a class. For our experiments we used a single multi-layer perceptron (mlp) architecture, although other architectures, such as a CNN or LSTM, are also possible.
\item Polyak coefficient: for stability, the network may only be allowed to change at a certain rate. A simple way to do this is by making a copy of the main network into a target network and averaging the parameters. The coefficient for Polyak-averaging indicates the weight for such an average.
\item Batch size: the batch size from the experience buffers that is used for training
\item Gamma: the discount gamma parameter value used for $Q$ learning updates
\item Q-network learning rate: the learning rate for the $Q$ value update function on the critic network
\item $\pi$-network learning rate: learning rate for the actor network
\item Normalization $\epsilon$:  epsilon used for observation normalization meant to avoid numerical instabilities
\item Normalization clipping: magnitude of normalized output is clipped to this value
\item Maximum action value: maximum magnitude for the each of the action coordinates
\item Action L2 penalty: quadratic penalty on the actions before rescaling by the maximum action value
\item Observation clipping: observation clipping before normalization
\item Size of rollout batch: refers to the number of parallel rollouts per DDPG agent/thread

\end{itemize}

\noindent We modified both the environment and the agent to record the instances when the robot grabs the box, the moment when it moves the box without grabbing it, and the (x,y) coordinates of the target location for each episode. To perform intra-episode analysis, we also made changes to collect and record separately the state space parameters for each step inside an episode.

\begin{figure}[h]
\centering
\includegraphics[width=0.98\linewidth]{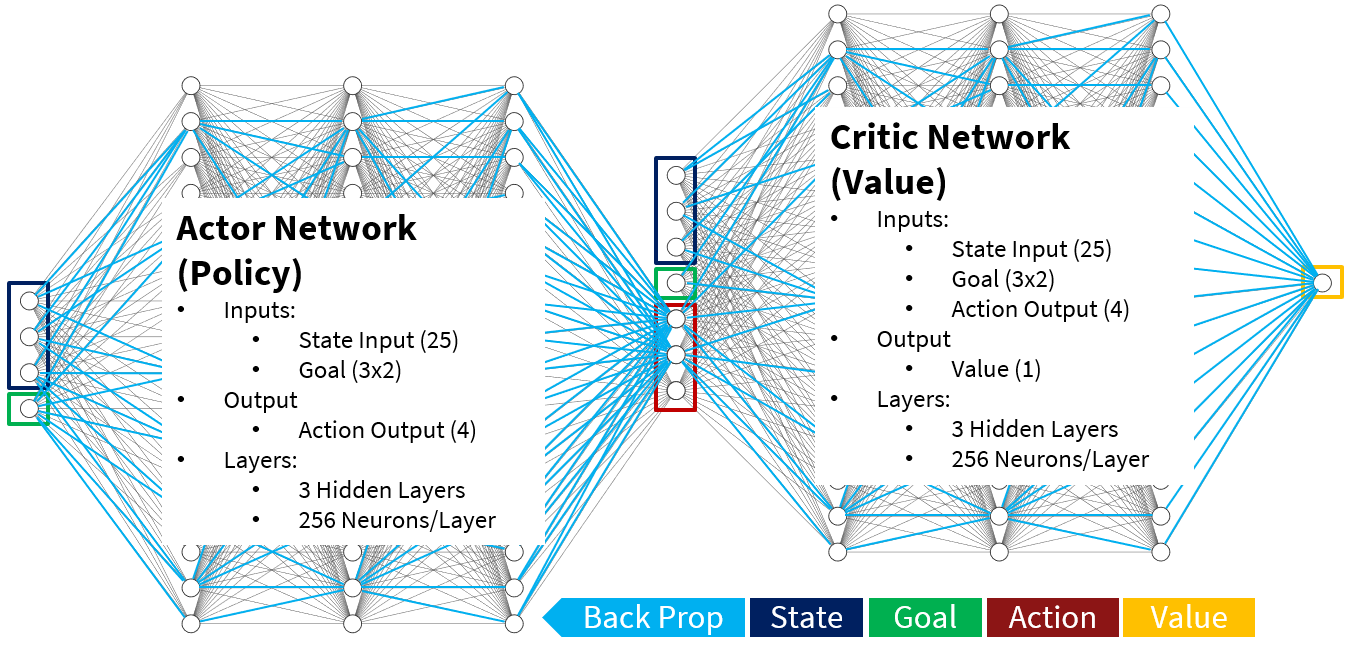}
\caption{Diagram of the neural network}
\label{fig:neural_network}
\end{figure}

\subsection*{Main Experiments}

\begin{table}[ht]
\begin{tabular}{cl}
\hline
\multicolumn{1}{l}{\textbf{Experiment}} & \textbf{Description}                                                                                                       \\ \hline
\multicolumn{1}{c|}{Ditch}                  & \begin{tabular}[c]{@{}l@{}}Target goal on the $40 \pm 15 cm$ range from the x axis, beyond the reach of the \\ hand manipulator. The region is separated by a short ditch 2 cm deep and 2.5 cm wide, \\ located at 26.25 cm from the origin.\\ \\ \end{tabular}                                                                                                             \\
\multicolumn{1}{c|}{Wall}                  & \begin{tabular}[c]{@{}l@{}}Target goal on the $40 \pm 15 cm$ range from the x axis, beyond the reach of the \\ hand manipulator. The region is separated by a short wall 3 cm high and 1 cm wide, \\ located 28 cm from the origin.\\ \\ \end{tabular}                                                                                               \\
\multicolumn{1}{c|}{TargetNear}                  & \begin{tabular}[c]{@{}l@{}}Target goal on the $40 \pm 30 cm$ range from the x axis, often but not always within \\ the reach of the hand manipulator. The region is separated by a short wall 3 cm high \\ and 1 cm wide, located 28 cm from the origin.\\ \\ \end{tabular}                                                                         \\
 \hline
\end{tabular}
\caption{Description of the main experiments}
\label{tab:experiments}
\end{table}

All experiments use the same \textit{Fetch} robot that was used by OpenAI \cite{DBLP:journals/corr/abs-1802-09464}, but the table has been modified to be bigger ($82.5cm$ long and $65.0cm$ wide) and include either the ditch or the wall constraint. The box has a mass of $2kg$. Table \ref{tab:experiments} contains additional details about each experiment setup.

\subsection*{Ancillary Experiments}

\begin{table}[ht]
\begin{tabular}{cl}
\hline
\multicolumn{1}{l}{\textbf{Experiment}} & \textbf{Description}                                                                                                       \\ \hline
\multicolumn{1}{c|}{TargetMoving}                  & \begin{tabular}[c]{@{}l@{}}Same as the \texttt{Wall} experiment, however the target region is moved backwards slowly \\ at a rate of $2.0 \times 10^{-6}cm$ per step.\\ \\ \end{tabular}                                                                                                \\
\multicolumn{1}{c|}{TargetExpanding}                  & \begin{tabular}[c]{@{}l@{}}Same as the \texttt{Wall} experiment, however the target region expands backwards and \\ sideways slowly at a rate of $6.67 \times 10^{-7}cm$ per step.\\ \\ \end{tabular}                                                                                                \\
\multicolumn{1}{c|}{RStateSp}                  & \begin{tabular}[c]{@{}l@{}}Same as the \texttt{Wall} experiment, however the box's orientation and rotational velocity \\
have been removed from the state space.\\ \\ \end{tabular}                                                                      \\
 \hline
\end{tabular}
\caption{Description of the ancillary experiments}
\label{tab:experiments_ancillary}
\end{table}

Three smaller experiments were created for ancillary purposes (Table \ref{tab:experiments_ancillary}). The \texttt{Wall-TargetMoving} and \texttt{Wall-TargetExpanding} environments were created to assess the impact of non-uniform distributions for target sampling. Both experiments use the pre-trained policy from the \texttt{Wall} experiment as a baseline. We correctly expected \texttt{Wall-TargetMoving} to perform very poorly as the experiment allows for a time dependence on the sampling, which is one of the more important issues that the experience buffer tries to solve. Moving the target area also breaks the uniform sampling. In this case, new information recently learned came to overwrite the old and performance deteriorates catastrophically. The \texttt{Wall-TargetExpanding} experiment performed better, as the uniform sampling was maintained despite the time dependence of the sampling. The \texttt{Wall-RStateSp} experiment repeated the \texttt{Wall} experiment from the beginning, removing the box's orientation and rotational velocity from the state space. Originally we anticipated little change, as the gripper cannot really rotate to pick up the box in a different orientation, and the starting orientation of the box is always aligned to the gripper's. However, the results show that learning did slow down when compared to the regular \texttt{Wall} experiment.

\end{document}